\documentclass{article}
\usepackage[utf8]{inputenc}
\usepackage{graphicx}
\usepackage{amsmath}
\usepackage{amsfonts}
\usepackage{amssymb}
\usepackage{nips15submit_e, times}
\usepackage[noend]{algorithm2e}
\nipsfinalcopy
\title{Parallel Markov Chain Monte Carlo for the Indian Buffet Process}
\author{
Michael M. Zhang\\
Department of Statistics and Data Science\\
The University of Texas at Austin\\
Austin, TX 78712\\
\texttt{michael\_zhang@utexas.edu}\\
\And
Avinava Dubey\\
Machine Learning Department\\
Carnegie Mellon University\\
Pittsburgh, PA 15213\\
\texttt{avinava.dubey@gmail.com}\\
\And
Sinead A. Williamson\\
McCombs School of Business\\
The University of Texas at Austin\\
Austin, TX 78712\\
\texttt{sinead.williamson@mccombs.utexas.edu}}
\begin{document}
\maketitle

\begin{abstract}
Indian Buffet Process based models are an elegant way for discovering underlying features within a data set, but inference in such models can be slow. Inferring underlying features using Markov chain Monte Carlo either relies on an uncollapsed representation, which leads to poor mixing, or on a collapsed representation, which leads to a quadratic increase in computational complexity. Existing attempts at distributing inference have introduced additional approximation within the inference procedure. In this paper we present a novel algorithm to perform asymptotically exact parallel Markov chain Monte Carlo inference for Indian Buffet Process models. We take advantage of the fact that the features are conditionally independent under the beta-Bernoulli process. 
Because of this conditional independence, we can partition the features into two parts: one part containing only the finitely many instantiated features and the other part containing the infinite tail of uninstantiated features. 
For the finite partition, parallel inference is simple given the instantiation of features. But for the infinite tail, performing uncollapsed MCMC leads to poor mixing and hence we collapse out the features. The resulting hybrid sampler, while being parallel, produces samples asymptotically from the true posterior.
\end{abstract}

\section{Introduction}
The Indian Buffet Process (IBP) \cite{ghahramani2005infinite} has proven to be a popular Bayesian non-parametric model for discovering underlying features within a data set. More explicitly, the IBP places a probability distribution over a binary matrix which indicates the presence of a particular feature $k$ for observation $n$. The non-parametric nature of the IBP means that we assume, \textit{a priori}, an infinite number of features, though only a finite number of those will be present \textit{a posteriori}.

The non-parametric nature of the IBP for feature discovery is very attractive in situations where the number of latent objects is unknown. However, more and more applications of Bayesian machine learning are focused towards ``big data'' settings, where the number of observations is extremely large.  If we consider an extremely large data set then the uncollapsed and collapsed samplers detailed in \cite{ghahramani2005infinite} will be ineffective--the uncollapsed sampler will mix poorly as the the dimensionality of the data increases, and the computational complexity of the collapsed sampler grows quadratically. \cite{doshi2009accelerated} presents an accelerated sampler for the IBP that exhibits the mixing quality of a collapsed sampler with the speed of an uncollapsed sampler.

If we were in a ``big data'' situation in which we cannot even hold our data on one machine, we then would need to perform parallel inference for the IBP. Comparatively fewer research exists on parallel inference for the IBP than for other popular Bayesian non-parametric models, like the Dirichlet Process mixture model \cite{ge2015distributed, williamson2013parallel}. \cite{broderick2012mad} and \cite{teh2007stick} offer \textit{parallelizable} solutions. But heretofore, only \cite{doshi2009large} has developed a parallel inference algorithm, and unlike the hybrid algorithm proposed in this paper, that algorithm is not asymptotically exact.

Our method is a hybrid Markov chain Monte Carlo (MCMC) algorithm that, asymptotically, will sample from the true posterior distribution. We exploit conditional independencies between features so that we may distribute inference for the binary matrix $Z$ across processors. Like the accelerated sampler of \cite{doshi2009accelerated}, our algorithm takes advantage of the speed of the uncollapsed algorithm along with the efficient mixing of the collapsed algorithm.
\section{Latent feature modeling using the Indian Buffet Process}\label{sec:IBP}
The IBP is a distribution over binary matrices $Z$ with infinitely many columns and exchangeable rows. This matrix can be used to select finite subsets of an unbounded set of latent features. For example, we can combine the IBP-distributed matrix $Z$ with a set $A=(A_k)$ of normally distributed features, $A_k \sim \mbox{Normal}(0, \sigma_A^2I)$, to get a latent feature model appropriate for real-valued images:
\begin{equation}
X = ZA+\epsilon \qquad \qquad \mbox{where }\epsilon \sim \mbox{Normal}(0, \sigma_X^2I). \label{eqn:gaussianllk}
\end{equation}
We can think of the IBP as arising out of the infinite limit, as $K\rightarrow \infty$, of an $N\times K$ matrix with entries generated according to
\begin{align}
\pi_k | \alpha \sim \text{Beta}\left( \frac{\alpha}{K}, 1 \right), \quad \quad \quad 
Z_{nk} | \pi_k \sim \text{Bernoulli}\left( \pi_k \right).\label{eqn:betabern}
\end{align}
Note that under this construction, the columns of $Z$ are independent. If we take $K\rightarrow \infty$ and marginalize out the beta random variables, we obtain a distribution over exchangeable matrices. We can describe the predictive distribution of this sequence using the following restaurant analogy: Customers enter a buffet restaurant with infinitely many dishes (representing columns). The first customer selects a $\mbox{Poisson}(\alpha)$ number of dishes (non-zero entries). The $n$th customer selects each previously-sampled dish $k$ with probability $m_k/n$, where $m_k $ is the number of customers who have already taken the $k$th dish. The $n$th customer then selects a $\mbox{Poisson}(\alpha/n)$ number of new dishes.

Both the beta-Bernoulli and the restaurant analogies suggest inference schemes. We can instantiate the $\pi_k$, either using a finite-dimensional approximation as in Equation~\ref{eqn:betabern}, or by using a truncation of the full infinite sequence as described by \cite{teh2007stick}. This approach is inherently parallelizable, since the rows of $Z$ are conditionally independent given the $\pi_k$. However, previously unseen features must be globally instantiated. As the dimensionality of our data grows, the chance of sampling a ``good'' previously unseen feature decreases, leading to slow mixing (as observed by \cite{doshi2009accelerated}).

Alternatively, we can integrate out the features and work in the collapsed restaurant representation. But this is difficult to parallelize, since the probability of selecting a feature depends on the global number of observations exhibiting that feature. Furthermore, each time a new feature is instantiated, that new feature must be communicated to all processors. The only existing parallel inference algorithm for the IBP uses approximations to avoid constantly updating the feature counts.
\section{The hybrid algorithm for parallel MCMC}
As we saw in Section~\ref{sec:IBP}, there are disadvantages to using both the collapsed and the uncollapsed representation, particularly in a distributed setting. We choose a third path, combining collapsed and uncollapsed methods in a hybrid approach. We note that the uncollapsed algorithm will generally perform well when working with popular features, where we are able to make use of information across processors about that feature's location. However, it will perform poorly when it comes to instantiating new features. Conversely, the collapsed algorithm performs well at introducing new features, but cannot be exactly parallelized without significant overhead.

We note that, as implied by Equation~\ref{eqn:betabern}, we can split $Z$ into two conditionally independent sub-matrices, one containing the first $K^{+}$ features, and the other containing the remaining features. We partition our features so that the currently instantiated features are in the first finite-dimensional matrix, and the infinite uninstantiated tail is in the second matrix. We perform uncollapsed sampling on the  $K^{+}$ instantiated features, and collapsed sampling to propose and sample the new $K^{\ast}$ features. We divide the $X$ and $Z$ matrices along the observation axis across $P$ processors and at each iteration, one processor, $p^{\prime}$, will be able to generate new features on the collapsed infinite tail, while all other processors perform uncollapsed inference restricted to using only the first $K^+$ features. Periodically, we will transfer newly instantiated features to the finite-dimensional subset of instantiated features, ensuring global consistency of the algorithm--in effect, the processor $p^{\prime}$ acts to propose new features to be added to the uncollapsed, finite-dimensional representation. 

A single iteration of the hybrid algorithm proceeds as follows:

\For{L sub-iterations}{
	\For{$p$ in $p=1,\ldots,P$}{
		\For{$n$ in $n=1,\ldots,N_p$}{	
                  \For{$k$ in $k=1,\ldots, K^+$}{
		    Sample $Z_{nk}$ according to
                    $$P(Z_{nk} = 1 | ...) \propto \pi_k P(X|Z, A)$$
                  }
                  \If{$p = p^{\prime}$}{
		    \For{ $k$ in $K^++1,\dots, K^{\ast}_{p}$}{
                      Sample $Z_{nk}$ according to
                      $$P(Z_{nk} = 1 | ...) \propto \frac{m_k - Z_{nk}}{N} \int \! P(X|Z,A^{+}, A^{\ast})P(A^{\ast}) \, \mathrm{d}A^{\ast}$$
		      Draw $K_{new} \sim \text{Poisson}\left( \alpha / N \right)$\\
		      Propose $K_{new}$ features from $P(K_{new}) \propto P(X | Z_{new})$, using a Metropolis-Hastings step\\
				}
			}
		}
	}
}	
\If{master processor}{
	Receive summary statistics from all other processors\\
	Update global counts of features, $m_k$	\\
	Sample posterior values for parameters $A$, $\sigma^2_X$, $\sigma^2_A$, $\pi_k$ and hyperparameter $\alpha$\\
	$K^{+} \leftarrow K^{+} + \sum_{p=1}^{P}K^{\ast}_p$\\
	$K \leftarrow K^{+}$\\
	$K^{\ast}_{p} \leftarrow 0$\\		
	Broadcast new parameters to all other processors\\
        Select $p^\prime\sim \mbox{Uniform}\{1,\dots,P\}$
}	

Note that
here we assume the linear Gaussian likelihood described in Equation \ref{eqn:gaussianllk}, but the hybrid algorithm can easily be adapted to use other likelihoods.
\section{Results}
\begin{figure}[hbtp]
\caption{$\log P(X,Z)$ over log time on held out test set}
\centering
\includegraphics[scale=.5]{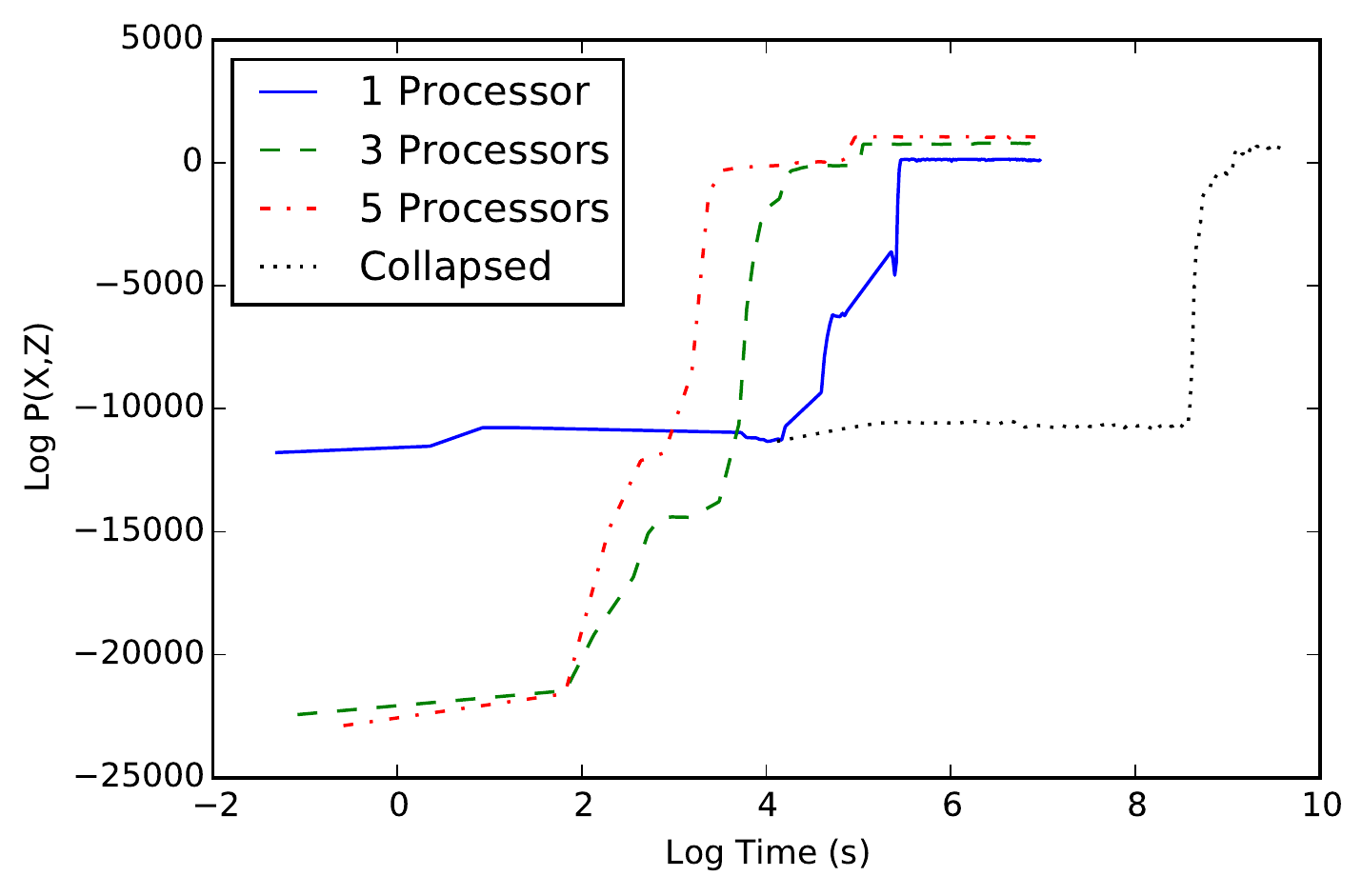}
\label{fig:lik}
\end{figure}

\begin{figure}[hbtp]
\caption{True features (top) and posterior features generated from the ``Cambridge" data set from the collapsed sampler (middle) and from the hybrid algorithm with 5 processors (bottom)}
\centering
\includegraphics[scale=.5]{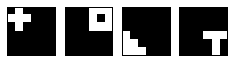}\\
\includegraphics[scale=.5]{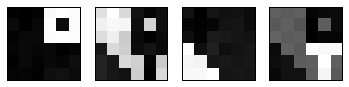}\\
\includegraphics[scale=.5]{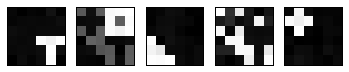}
\label{fig:features}
\end{figure}

To evaluate our algorithm, we calculate the joint log likelihood of $P(X,Z)$ on a held-out evaluation set of data and monitor the joint likelihood over log time against the collapsed sampler for the IBP. The data used for our evaluation is the $1000 \times 36$ dimension canonical ``Cambridge'' synthetic data set seen in \cite{griffiths2011indian}. We ran the hybrid algorithm on 1, 3, and 5 processors for 1000 iterations and 5 sub-iterations per global step on code written in Python. The data and the MCMC inference were distributed across processors through Message Passing Interface using \texttt{mpi4py}. As seen in Figure~\ref{fig:lik}, we can see that adding additional processors gives significant speedup, without a big difference in estimate quality. Interestingly, even with one processor, our algorithm converges faster than a purely collapsed sampler.   The resulting posterior features are seen in Figure~\ref{fig:features}.
\section{Conclusion}
We presented in this paper an asymptotically exact parallel Markov chain Monte Carlo algorithm for inference in an Indian Buffet Process feature model. Our parallelization technique exploits the independence between observations for the uncollapsed sampling of the feature indicators while using a partially collapsed sampler to infer values of $Z$ for new features. However, we still may face significant overhead in sending summary statistics to the master processor and broadcasting the new posterior draws for the parameters to the other processors. Directions for future research may point towards developing clever techniques to reduce this possible computational bottleneck. Regardless, our novel algorithm pushes the IBP closer towards high scalability. In comparison, previous research in parallel inference algorithms for IBP models is limited to \cite{doshi2009large}, who present an exact Metropolis-Hastings sampler but, in practice, use an approximate sampler. 

Parallelization is important in settings where we may have a huge number of observations. Direct implementation of the IBP with earlier inference algorithms in a ``big data'' scenario on a single machine will undoubtedly lead to poor results or inefficient computing. The algorithm in this paper avoids these problems and is guaranteed to produce results exact to the non-parallel inference method. ``Big data'' is an increasingly important concern for machine learning tasks because the nature of the data available now has grown to such a massive size that the scalability of an algorithm needs to be a primary concern in developing machine learning tools. Inference in the IBP has generally been difficult but we have developed an inference algorithm that has made the IBP amenable to larger data sets.
\bibliographystyle{plain}
\bibliography{parallel_ibp}
\end{document}